\pdfoutput=1
\documentclass[11pt]{article}

\usepackage{acl}

\usepackage{times}
\usepackage{latexsym}

\usepackage{multirow}
\usepackage{graphicx}
\usepackage{subfigure}

\usepackage[T1]{fontenc}
\usepackage[utf8]{inputenc}

\usepackage{microtype}

\title{MultiWOZ 2.4: A Multi-Domain Task-Oriented Dialogue Dataset with Essential Annotation Corrections to Improve State Tracking Evaluation}

\author{Fanghua Ye \\
  University College London \\
  London, UK \\
  \\\And
  Jarana Manotumruksa \\
  University College London \\
  London, UK \\
  \texttt{\{fanghua.ye.19, j.manotumruksa, emine.yilmaz\}@ucl.ac.uk} \\\And
  Emine Yilmaz \\
  University College London \\
  London, UK \\
  \\}

\begin{document}
\maketitle
\begin{abstract}
The MultiWOZ 2.0 dataset has greatly stimulated the research of task-oriented dialogue systems. However, its state annotations contain substantial noise, which hinders a proper evaluation of model performance. To address this issue, massive efforts were devoted to correcting the annotations. Three improved versions (i.e., MultiWOZ 2.1-2.3) have then been released. Nonetheless, there are still plenty of incorrect and inconsistent annotations. This work introduces MultiWOZ 2.4, which refines the annotations in the validation set and test set of MultiWOZ 2.1. The annotations in the training set remain unchanged (same as MultiWOZ 2.1) to elicit robust and noise-resilient model training. We benchmark eight state-of-the-art dialogue state tracking models on MultiWOZ 2.4. All of them demonstrate much higher performance than on MultiWOZ 2.1\footnote{MultiWOZ 2.4 is released to the public at \url{https://github.com/smartyfh/MultiWOZ2.4}.}.
\end{abstract}

\section{Introduction}

In recent years, tremendous advances have been made in the research of task-oriented dialogue systems, attributed to a number of publicly available dialogue datasets like DSTC2~\citep{henderson-etal-2014-second}, FRAMES~\citep{el-asri-etal-2017-frames},  WOZ~\citep{wen-etal-2017-network}, M2M~\citep{shah-etal-2018-bootstrapping}, MultiWOZ 2.0~\citep{budzianowski-etal-2018-multiwoz}, SGD~\citep{rastogi2020towards}, CrossWOZ~\citep{zhu-etal-2020-crosswoz}, RiSAWOZ~\citep{quan-etal-2020-risawoz}, and TreeDST~\citep{cheng-etal-2020-conversational}. Among them, MultiWOZ 2.0 is the first large-scale dataset spanning multiple domains and thus has attracted the most attention. 

However, substantial noise has been found in the dialogue state annotations of MultiWOZ 2.0~\citep{eric-etal-2020-multiwoz}. To remedy this issue, \citet{eric-etal-2020-multiwoz} fixed $32\%$ of dialogue state annotations across $40\%$ of the dialogue turns, resulting in an improved version MultiWOZ 2.1. Despite the significant improvement in annotation quality, MultiWOZ 2.1 still severely suffers from incorrect and inconsistent annotations~\citep{zhang-etal-2020-find, hosseini2020simple}. The state-of-the-art joint goal accuracy~\citep{zhong-etal-2018-global} for dialogue state tracking on MultiWOZ 2.1 is merely around $60\%$~\citep{li2020coco}. Even worse, the noise in the validation set and test set makes it relatively challenging to assess model performance properly and adequately. To reduce the impact of noise, different preprocessing strategies have been utilized by existing models. For example, TRADE~\citep{wu-etal-2019-transferable} fixes some general annotation errors. SimpleTOD~\citep{hosseini2020simple} cleans partial noisy annotations in the test set. TripPy~\citep{heck-etal-2020-trippy} constructs a label map to handle value variants. These preprocessing strategies, albeit helpful, lead to an unfair performance comparison.

\begin{figure*}[ht]
  \centering
  \includegraphics[width=\linewidth]{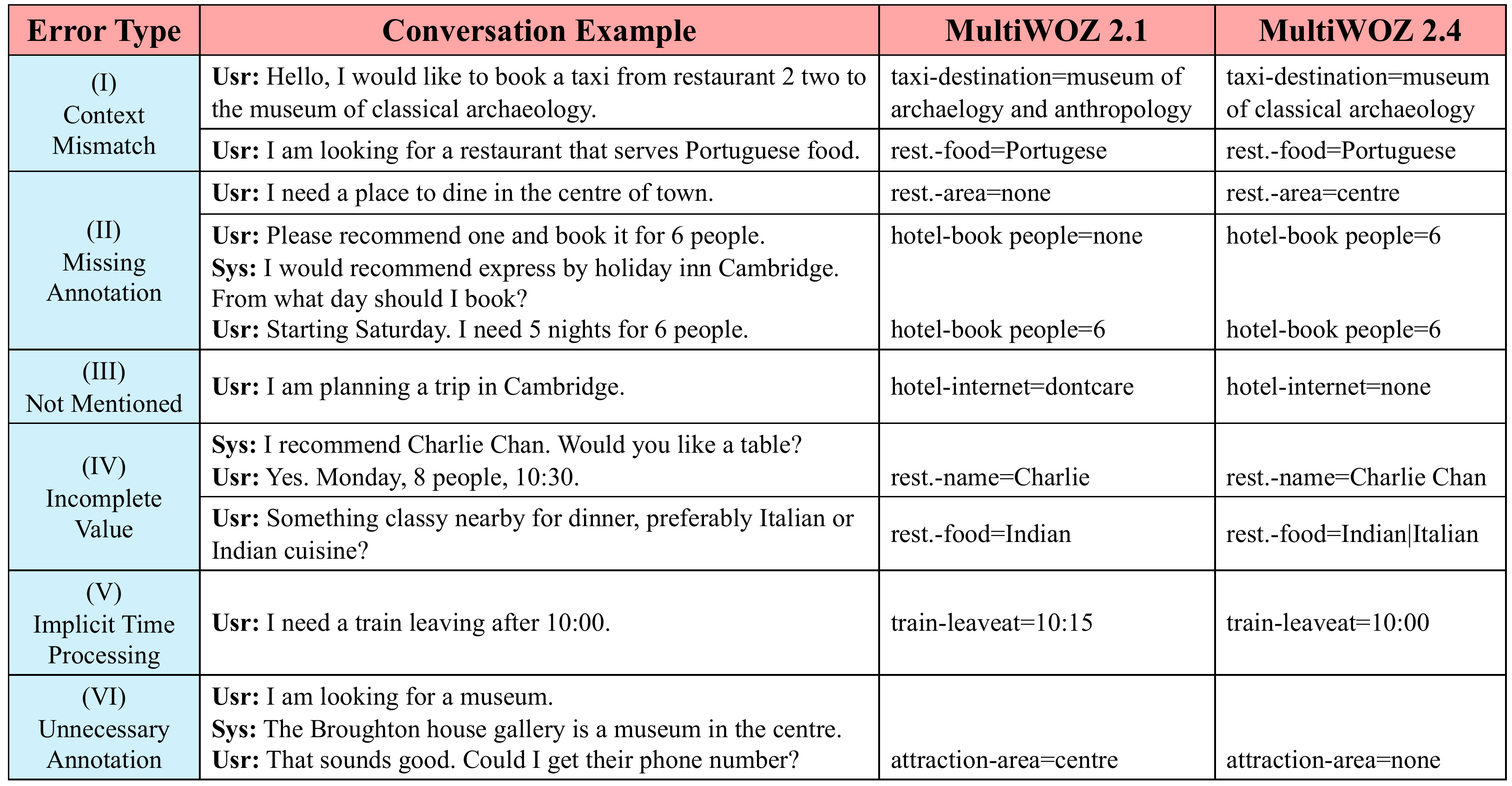}
  \vspace{-0.7cm}
  \caption{Examples of each error type.  Only the problematic slots are presented. ``rest.'' is short for restaurant.}
  \label{fig:example}
 \vspace{-0.15cm}
\end{figure*}

Massive efforts have been made to further improve the annotation quality of MultiWOZ 2.1, resulting in MultiWOZ 2.2~\citep{zang-etal-2020-multiwoz} and MultiWOZ 2.3~\citep{han2020multiwoz}. However, they both have some limitations. More concretely, MultiWOZ 2.2 allows the presence of multiple values in the dialogue state. But it does not cover all the value variants. This incompleteness brings about serious inconsistencies. MultiWOZ 2.3 focuses on dialogue act annotations. The noise on dialogue state annotations has not been fully resolved.

In this work, we introduce MultiWOZ 2.4, an updated version on top of MultiWOZ 2.1, to improve dialogue state tracking evaluation. Specifically, we identify incorrect and inconsistent annotations in the validation set and test set, and fix them meticulously. This refinement results in changes to the state annotations of more than $41\%$ of turns over $65\%$ of dialogues. Since our main purpose is to improve the correctness and fairness of model evaluation, the annotations in the training set remain unchanged. Even so, our empirical study shows that much better performance can be achieved on MultiWOZ 2.4 than on all the previous versions. Furthermore, a noisy training set motivates us to design robust and noise-resilient training mechanisms, e.g., data augmentation~\citep{summerville-etal-2020-tame} and noisy label learning~\citep{han2020survey}. Considering that collecting noise-free large multi-domain dialogue datasets is costly and labor-intensive, we believe that training robust dialogue state tracking models from noisy training data will be of great interest to both industry and academia.

% (i.e., MultiWOZ 2.0-2.3)

\section{Annotation Refinement}

In MultiWOZ 2.0 \& 2.1, the dialogue state is represented as a series of \textit{slot-value} pairs. For example, \textit{attraction-area=centre} means that the slot is \textit{attraction-area} and its value is \textit{centre}. Considering that MultiWOZ 2.1 has significantly improved the annotation quality of MultiWOZ 2.0, we choose to continue the refinement on the basis of MultiWOZ 2.1. Another choice is to perform the refinement on top of MultiWOZ 2.2. However, as mentioned earlier, MultiWOZ 2.2 allows each slot to have multiple value variants. This relaxation increases the difficulty of annotating. It is challenging to include all the value variants. New value variants may also emerge as time goes by. Even worse, some value variants are ambiguous and invalid. For instance, ``Peking'' can be a shared variant of ``Peking University'' and ``Peking restaurant''. Hence, it is an ambiguous value variant. Besides, the benchmark evaluation on MultiWOZ 2.2 shows no evident performance improvements over MultiWOZ 2.1 \cite{zang-etal-2020-multiwoz}. In light of these, MultiWOZ 2.1 is a better basis for our refinement. 

% Since a dialogue may involve multiple domains and each domain also has multiple slots, it is impractical to ensure that the state annotations obtained via a crowdsourcing process are consistent and noise-free. 

% 

\subsection{Annotation Error Types}

The main goal of dialogue state tracking is to track what has been uttered by a user. Thus, it is generally assumed that the dialogue state should mainly rely on user utterances\footnote{If the user requirements cannot be satisfied (e.g., a restaurant asked by the user does not exist), the system should still track the ``wrong'' requirements as the dialogue state and then ask a clarification question \cite{dougan2022asking} to the user.}. Based on this assumption, we identify and fix six types of annotation errors in the validation set and test set of MultiWOZ 2.1. Figure~\ref{fig:example} shows examples for each error type.

\noindent \textbf{Context Mismatch:} The slot value is inconsistent with the one mentioned in the dialogue context. We also include values with typos in this error type.
    
\noindent \textbf{Missing Annotation:} The slot is  unlabelled, even though its value has been mentioned. In some cases, the annotations are delayed to later turns.
    
\noindent \textbf{Not Mentioned:} The slot has been annotated, however, its value has not been mentioned at all.
    
\noindent \textbf{Incomplete Value:} The slot value is a substring or an abbreviation of its full shape (e.g., ``Thurs'' vs. ``Thursday''). In some cases, the slot should have multiple values, but not all values are included.

\noindent \textbf{Implicit Time Processing:} This relates to the slots that take time as the value. Instead of copying the time specified in the dialogue context, the value has been implicitly processed (e.g., adding 15 min)\footnote{The value is implicitly processed when the time is after or before a certain point. Albeit reasonable, it is hard to decide the exact time offset. Thus, we copy the specified time directly.}.

% These annotations are not wrong, but they are    
\noindent \textbf{Unnecessary Annotation:} These unnecessary annotations exacerbate inconsistencies as different annotators have different opinions on whether to annotate these slots or not. In general, the values of these slots are mentioned by the system to respond to previous user requests or provide supplementary information. We found that in most dialogues, these slots are not annotated. Hence, we remove these annotations. However, the \verb|name|-related slots are an exception. If the user requests more information (e.g., \textit{address} and \textit{postcode}) about the recommended ``name'', the slots will be annotated.

\begin{table}[t]
\centering
\setlength{\tabcolsep}{2.0pt}
\begin{tabular}{l|cc}
\hline
\textbf{Refinement Type}   & \textbf{Count}  & \textbf{Ratio($\%$)} \\ \hline
no change           & 432,972 & 97.90 \\
\verb|none|$\rightarrow$value   & 3,230   & 0.73  \\
valueA/\verb|dontcare|$\rightarrow$valueB       & 1,598   & 0.36  \\
value/\verb|dontcare|$\rightarrow$\verb|none|          & 2,846   & 0.64  \\
\verb|none|/value$\rightarrow$\verb|dontcare| & 1,614   & 0.36  \\ \hline
\end{tabular}
\vspace*{-0.2cm}
\caption{The count and ratio of slot values changed in MultiWOZ 2.4 compared with MultiWOZ 2.1.}
\label{tab:countandratio}
\vspace*{-0.345cm}
\end{table}

\subsection{Annotation Refinement Procedure}
\label{sec:procedure}

The validation set and test set of MutliWOZ 2.1 contain 2,000 dialogues with more than 14,000 dialogue turns. These dialogues span over 5 domains with a total of 30 slots. To guarantee that the refined annotations are as correct and consistent as possible, we decided to rectify the annotations by ourselves rather than crowd-workers. However, if we check the annotations of all 30 slots at each turn, the workload is too heavy. To ease the burden, we instead only checked the annotations of turn-active slots. A slot being turn-active means that its value is determined by the dialogue context of current turn and is not inherited from previous turns. The average number of turn-active slots in the original annotations and in the refined annotations is 1.16 and 1.18, respectively. The full dialogue state is then obtained by accumulating all turn-active states from the first turn to current turn.

We also observed that some slot values are mentioned in different forms, such as ``concert hall'' vs. ``concerthall'' and ``guest house'' vs. ``guest houses''. The \verb|name|-related slot values may have a word \verb|the| at the beginning, e.g., ``Peking restaurant'' vs. ``the Peking restaurant''. We normalized these variants by selecting the one with the highest frequency. In addition, all \verb|time|-related slot values have been updated to the 24:00 format. We performed the above refining process twice to reduce mistakes and it took us one month to finish this task.

\subsection{Statistics on Refined Annotations}

Table~\ref{tab:countandratio} shows the count and percentage of slot values changed in MultiWOZ 2.4 compared with MultiWOZ 2.1. Note that \verb|none| and \verb|dontcare| are regarded as two special values. As can be seen, most slot values remain unchanged. This is because a dialogue only has a few active slots and the other slots always take the value \verb|none|. Table~\ref{tab:dataratio} further reports the ratio of refined slots, turns and dialogues. Here, the ratio of refined slots is computed on the basis of refined turns. It is shown that the corrected states relate to more than $41\%$ of turns over $65\%$ of dialogues. On average, the annotations of 1.53 ($30 \times 5.10\%$) slots at each refined turn have been rectified.

\begin{table}[t]
\centering
\setlength{\tabcolsep}{5pt}
\begin{tabular}{l|ccc}
\hline
\textbf{Dataset}  & \textbf{Slot($\%$)} & \textbf{Turn($\%$)}  & \textbf{Dialogue($\%$)} \\ \hline
val   & 5.04 & 42.61 & 67.40    \\
test  & 5.17 & 39.74 & 64.16    \\ \hline
total & 5.10 & 41.17 & 65.78    \\ \hline
\end{tabular}
\vspace*{-0.2cm}
\caption{The ratio of refined slots, turns and dialogues.}
\label{tab:dataratio}
\vspace*{-0.3cm}
\end{table}

\begin{figure}[t]
  \centering
  \includegraphics[width=0.92\linewidth]{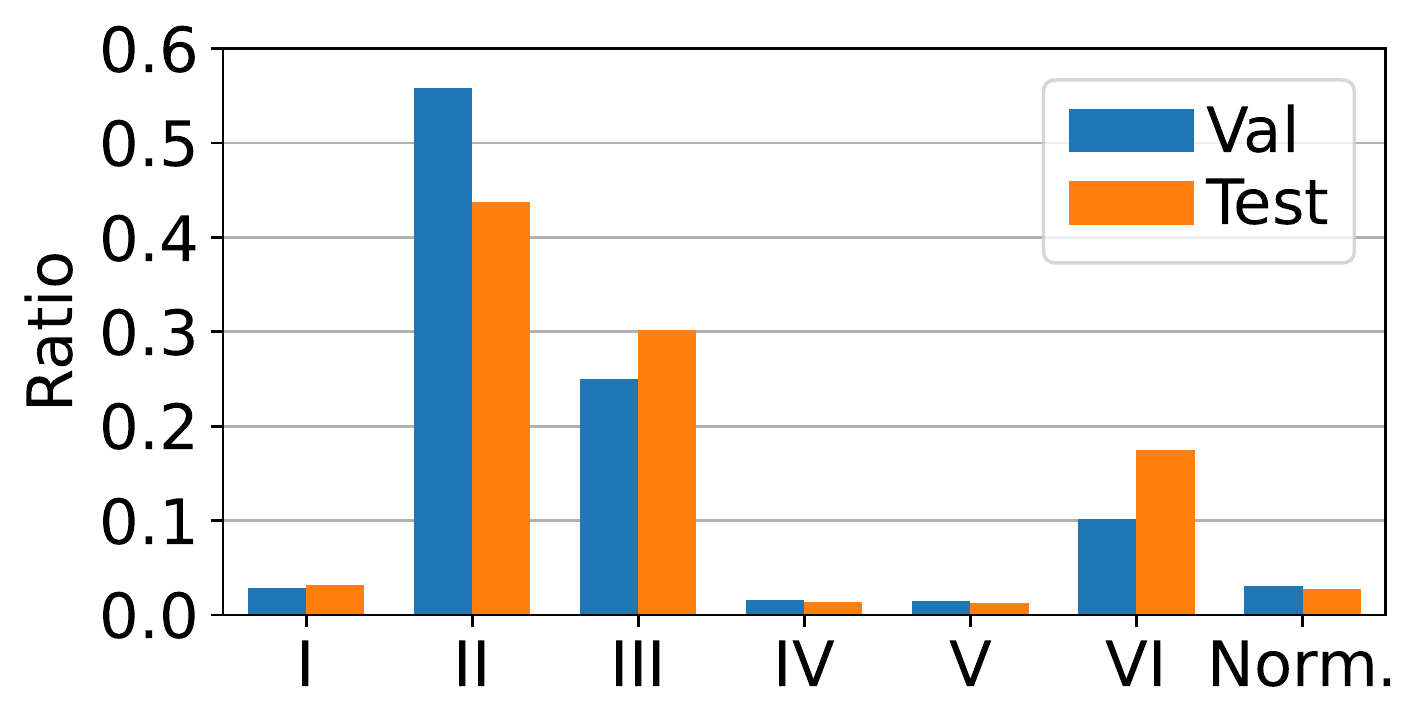}
  \vspace{-0.3cm}
  \caption{The ratio of different error types. ``Norm.'' refers to values normalized based on their frequency.}
  \label{fig:error-dist}
  \vspace{-0.3cm}
\end{figure}

Figure~\ref{fig:error-dist} illustrates the distribution of different error types. We also treat unnormalized values (cf. \S\ref{sec:procedure}) as a special type of errors. Figure~\ref{fig:error-dist} shows that ``\textit{Missing Annotation}'' and ``\textit{Not Mentioned}'' are the two most frequent error types. It also shows that more than $10\%$ of errors are related to ``\textit{Unnecessary Annotation}'', while the other types of errors only account for a relatively small proportion.

% \usepackage{multirow}
\begin{table*}[t]
\centering
% \setlength{\tabcolsep}{4.0pt}
% \small
\fontsize{10.7}{12}\selectfont
\begin{tabular}{c|c|ccc|cc}
\hline
\multirow{2}{*}{}                                                              & \multirow{2}{*}[-0.57em]{\textbf{Model}} & \multicolumn{3}{c|}{\textbf{Joint Goal Accuracy ($\%$)}}                                                                                                                                                & \multicolumn{2}{c}{\textbf{Slot Accuracy ($\%$)}}                                                                                         \\ \cline{3-7} 
                                                                               &                        & \begin{tabular}[c]{@{}c@{}}\textbf{MWZ 2.1} \\ \textbf{Test}\end{tabular} & \begin{tabular}[c]{@{}c@{}}\textbf{MWZ 2.4}\\ \textbf{Test}\end{tabular} & \begin{tabular}[c]{@{}c@{}}\textbf{MWZ 2.4}\\ \textbf{Val}\end{tabular} & \begin{tabular}[c]{@{}c@{}}\textbf{MWZ 2.1}\\ \textbf{Test}\end{tabular} & \begin{tabular}[c]{@{}c@{}}\textbf{MWZ 2.4}\\ \textbf{Test}\end{tabular} \\ \hline
\multirow{2}{*}{\begin{tabular}[c]{@{}c@{}}predefined\\ ontology\end{tabular}} & SUMBT                  & 49.01                                                        & 61.86 \small{(+12.85)}                                              & 62.31                                                      & 96.76                                                       & 97.90                                                       \\
                                                                            %   & CHAN                   & 53.38                                                        & 68.25 \small{(+14.87)}                                              & 68.23                                                      & 97.39                                                       & 98.52                                                       \\
                                                                               & STAR                   & \textbf{56.36}                                                        & \textbf{73.62} \small{(+17.26)}                                              & 74.59                                                      & \textbf{97.59}                                                       & \textbf{98.85}                                                       \\ \hline
\multirow{6}{*}{\begin{tabular}[c]{@{}c@{}}open \\ vocabulary\end{tabular}}    & TRADE                  & 45.60                                                        & 55.05 \small{(+9.45)}                                               & 57.01                                                      & 96.55                                                       & 97.62                                                       \\
                                                                               & PIN                    & 48.40                                                        & 58.92 \small{(+10.52)}                                              & 60.37                                                      & 97.02                                                       & 98.02                                                       \\
                                                                               & SOM-DST                & 51.24                                                        & \textbf{66.78} \small{(+15.54)}                                              & 68.77                                                      & 97.15                                                       & \textbf{98.38}                                                       \\
                                          & SimpleTOD & 51.75 & 57.18 \small{(+5.43)} & 55.02 & 96.78 & 96.97 \\
                                                                               & SAVN                   & 54.86                                                        & 60.55 \small{(+5.69)}                                               & 61.91                                                      & \textbf{97.55}                                                       & 98.05                                                       \\
                                                                               & TripPy                 & \textbf{55.18}                                                        & 64.75 \small{(+9.57)}                                               & 64.27                                                      & 97.48                                                       & 98.33                                                       \\ \hline
\end{tabular}
\vspace*{-0.25cm}
\caption{Joint goal accuracy and slot accuracy of different models on MultiWOZ 2.1 and MultiWOZ 2.4. }
\label{tab:baselinecomp}
\vspace*{-0.3cm}
\end{table*}

\begin{table}[t]
\centering
\fontsize{10.7}{12}\selectfont
\setlength{\tabcolsep}{3.3pt}
\begin{tabular}{l|cc}
\hline
\textbf{Dataset }         & \textbf{SUMBT ($\%$)} & \textbf{TRADE ($\%$)} \\ \hline
MultiWOZ 2.0     & 48.81 & 48.62  \\
MultiWOZ 2.1     & 49.01 & 45.60 \\
MultiWOZ 2.2     & 49.70 & 46.60 \\
MultiWOZ 2.3     & 52.90 & 49.20 \\
MultiWOZ 2.3-cof & 54.60 & 49.90 \\ \hline
MultiWOZ 2.4     & \textbf{61.86} & \textbf{55.05} \\ \hline
\end{tabular}
\vspace*{-0.2cm}
\caption{Comparison of test set joint goal accuracy on different versions of the MultiWOZ dataset.}
\label{tab:versionscomp}
\vspace*{-0.3cm}
\end{table}

\section{Benchmark Evaluation}
\label{sec:bbbb}

\subsection{Benchmark Models}

Existing neural dialogue state tracking models can be roughly divided into two categories: predefined ontology-based methods and open vocabulary-based methods. The ontology-based methods perform classification by scoring all possible slot-value pairs in the ontology and selecting the value with the highest score as the prediction. By contrast, the open vocabulary-based methods directly generate or extract slot values from the dialogue context. We benchmark the performance of our refined dataset on both types of methods, including SUMBT~\citep{lee-etal-2019-sumbt}, STAR~\citep{ye2021slot}, TRADE~\citep{wu-etal-2019-transferable}, PIN~\citep{chen-etal-2020-parallel}, SOM-DST~\citep{kim-etal-2020-efficient}, SimpleTOD~\citep{hosseini2020simple}, SAVN~\citep{wang-etal-2020-slot}, and TripPy~\citep{heck-etal-2020-trippy}.

\subsection{Benchmark Results}

We adopt joint goal accuracy~\citep{zhong-etal-2018-global} and slot accuracy as evaluation metrics. The joint goal accuracy is defined as the ratio of dialogue turns in which all slot values are correctly predicted. The slot accuracy is defined as the average accuracy of all slots. As shown in Table~\ref{tab:baselinecomp}, all models achieve much higher performance on MultiWOZ 2.4.  SimpleTOD shows the least performance improvement. The reason may be that SimpleTOD generates state values directly while other methods such as TRADE leverage the copy mechanism \citep{see-etal-2017-get} to assist in the generation process. SAVN also shows a low performance increase, as it has already utilized value normalization to tackle label variants in MultiWOZ 2.1. 
We then report the joint goal accuracy of SUMBT and TRADE on different versions of the dataset in Table~\ref{tab:versionscomp}, in which MultiWOZ 2.3-cof means MultiWOZ 2.3 with co-reference applied. As can be seen, both methods perform better on MultiWOZ 2.4 than on all previous versions. We include the domain-specific accuracy of SOM-DST and STAR in Table~\ref{tab:domaincomp}, which shows that except SOM-DST in the \textit{taxi} domain, both methods demonstrate higher performance in each domain of MultiWOZ 2.4.

% The ontology-based models demonstrate the highest performance promotion, which benefits from the improved ontology.

% 

% \usepackage{multirow}
\begin{table}[t]
\centering
\fontsize{10.7}{12}\selectfont
\begin{tabular}{l|cc|cc}
\hline
\multirow{2}{*}{\textbf{Domain}} & \multicolumn{2}{c|}{\textbf{SOM-DST ($\%$)}} & \multicolumn{2}{c}{\textbf{STAR ($\%$)}} \\ \cline{2-5} 
                        & \textbf{2.1}       & \textbf{2.4}      & \textbf{2.1}     & \textbf{2.4}    \\ \hline
attraction              & 69.83         & 83.22        & 70.95       & 84.45      \\
hotel                   & 49.53         & 64.52        & 52.99       & 69.10      \\
restaurant              & 65.72         & 77.67        & 69.17       & 84.20      \\
taxi                    & 59.96         & 54.76        & 66.67       & 73.63      \\
train                   & 70.36         & 82.73        & 75.10       & 90.36      \\ \hline
\end{tabular}
\vspace*{-0.2cm}
\caption{Comparison of domain-specific test set joint goal accuracy.}
\label{tab:domaincomp}
\vspace*{-0.415cm}
\end{table}

\section{Human Evaluation}

We also perform a human evaluation on the quality of the refined annotations. We randomly sampled 50 dialogues from the test set and recruited 5 computer science students to compare our refinement against the annotations in MultiWOZ 2.1. Specifically, the raters were asked to assign a score to each turn of the sampled dialogues based on the following criteria: 1) \textbf{-2}: A score of -2 means that both the refined annotation and original annotation are not completely correct; 2) \textbf{-1}: A score of -1 means that the original annotation is correct while the refined annotation is problematic; 3) \textbf{0}: A score of 0 means that both the refined annotation and original annotation are correct, that is, no changes have been made to the original annotation; 4) \textbf{1}: A score of 1 means that the refined annotation is correct while the original annotation is invalid. 

We obtain an average score of $0.1653$, meaning that our refined annotations are more accurate. We further employ Fleiss' kappa~\citep{fleiss1971measuring} to measure the level of agreement among different raters. We obtain $\kappa=0.9226$, which indicates an almost perfect agreement across the five raters.

% \section{Score Distribution of Human Evaluation}
% \label{sec:hv_dsit}

\begin{figure}[t]
  \centering
  \includegraphics[width=1\linewidth]{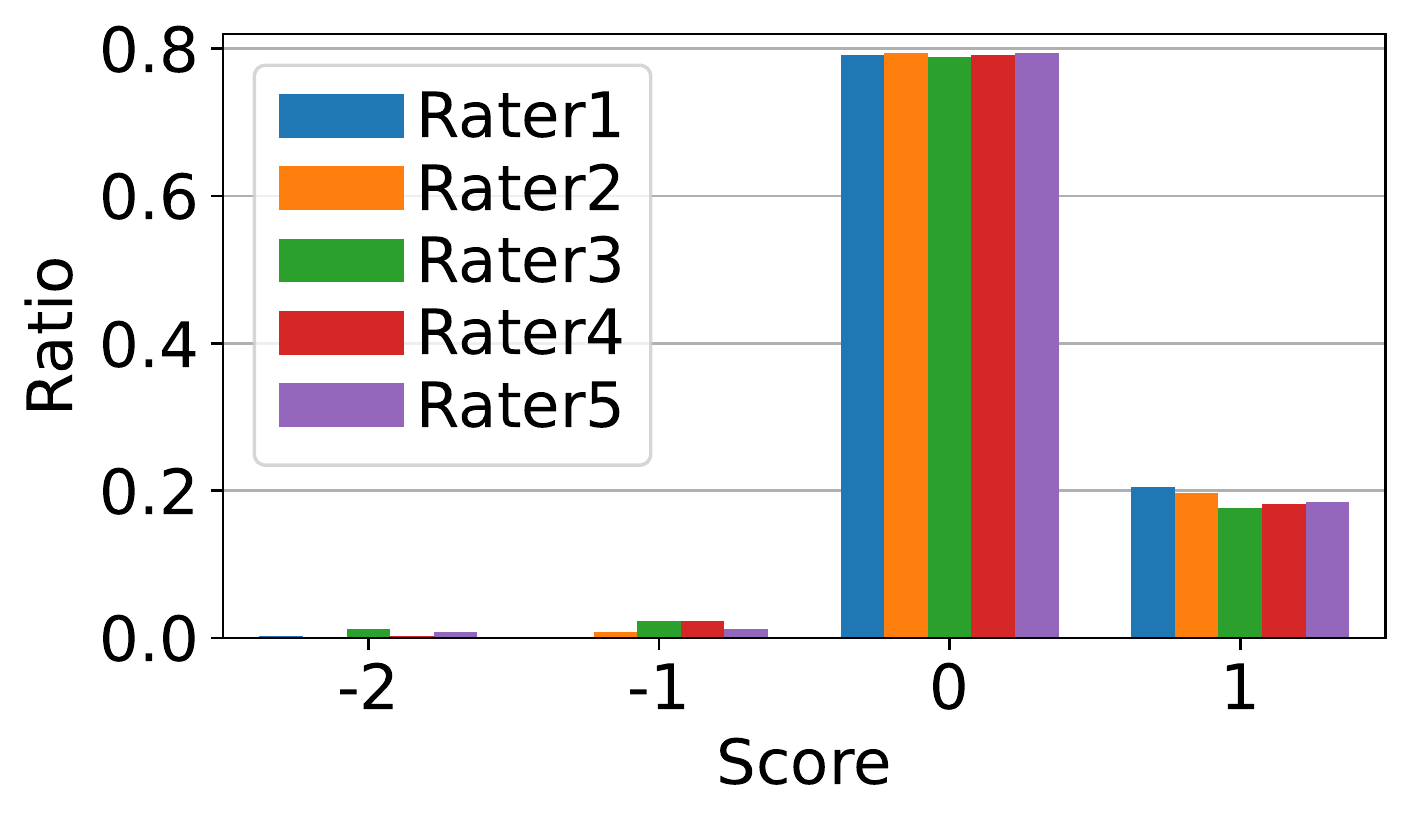}
  \vspace{-0.8cm}
  \caption{The score distribution of different raters.}
  \label{fig:score-dist}
  \vspace{-0.3cm}
\end{figure}

We illustrate the score distributions of different raters in Figure~\ref{fig:score-dist}. From this figure, we can intuitively observe that there is a high level of agreement among the five raters. Figure~\ref{fig:score-dist} also shows that in most cases, the refined annotation and the original annotation are both correct, meaning that there is no need to make any changes to the original annotation. This is desirable, as our refinement is based on MultiWOZ 2.1 which has already fixed lots of annotation errors. Around $20\%$ of annotations in MultiWOZ 2.4 are deemed to be more accurate than MultiWOZ 2.1, while only about $1\%$ of annotations in MultiWOZ 2.1 are evaluated as better. This verifies again that our refinement has higher quality.

We further inspected the annotations in MultiWOZ 2.1 that are assessed to be more appropriate. We found that these annotations are mainly related to the slot \textit{hotel-type}. This slot has four candidate values \{``hotel'', ``guest house'', ``none'', ``dontcare''\}, which are relatively confusing because the term ``hotel'' is also one candidate value. In practice, when a user says ``I am looking for a hotel with 4 stars'', the user may actually mean that ``I am looking for a place to stay with 4 stars''. However, by convention, the term ``hotel'' is used more often, even though the user does not mean that the hotel type must be ``hotel''. In our refinement procedure, we chose to annotate this slot based on the whole dialogue session to understand the true user intention (i.e., \textit{hotel type=hotel}?) while the raters tended to take into account only the dialogue history. This ambiguous slot tells us that it is crucial to develop appropriate slots and candidate values that will not cause any confusions to the annotators.

\section{Caveats and Lessons Learned}

Although we have tried our best to correct as many annotations in the validation set and test set as possible, it is unlikely that we have fixed all the annotation errors. In fact, there are several challenges we faced during the refinement process that are particularly difficult to overcome. Firstly, as discussed earlier, the candidate values of some slots are confusing, which makes it really challenging to choose the most appropriate value. Secondly, in some scenarios, the user intention can have different interpretations. For example, the user utterance ``the hotel does not need to have internet though'' can mean that the user does not need internet at all (\textit{hotel-internet=no}) or the user does not care about if the internet is provided (\textit{hotel-internet=dontcare}). Thirdly, some slots may have multiple values. Sometimes these values should even be ordered according to users' preferences. When there are too many values (more than two), it is also questionable if the corresponding slot should be annotated. Suppose that the system recommended 10 museums to the user and the user asked ``Does any of them have zero entrance fee?'', should the slot \textit{attraction-name} be annotated?

Further, the dialogue state can be regarded as a structured representation of the complex user intentions. Due to the complexity of the language itself, some information will be inevitably lost when transforming unstructured user utterances into structured state representations. In this regard, dialogue state annotating is in essence a challenging task.

Given these challenges, it is necessary to define unambiguous slots and unconfusable candidate values to facilitate state annotating. It is also important to provide annotators with full instructions for each slot so that they can make consistent annotations.

% Nonetheless, we should still be able to collect high-quality albeit not noise-free annotations as long as proper instructions are provided to the annotators. 

\section{Conclusion}
% improve dialogue state tracking evaluation. 

We introduce MultiWOZ 2.4, an updated version of MultiWOZ 2.1, by rectifying (almost) all the annotation errors in the validation set and test set. We keep the annotations in the training set as is to encourage robust and noise-resilient model training. We further benchmark eight state-of-the-art dialogue state tracking models on MultiWOZ 2.4 to facilitate future research. All the benchmark models have demonstrated much better performance on MultiWOZ 2.4 than on MultiWOZ 2.1.

% verifying the quality of our refined annotations.

% Please check the Appendices for more analyses.

MultiWOZ 2.4 can also be applied to train better overall dialogue systems, e.g, by utilizing data augmentation techniques to generate high-quality training data based on the clean validation set.

\section*{Acknowledgments}
This project was funded by the EPSRC Fellowship titled “Task Based Information Retrieval” and grant reference number EP/P024289/1.

\bibliography{anthology,custom}
\bibliographystyle{acl_natbib}

\clearpage
\appendix

\section{Additional Statistics on the Refined Annotations}
In Table~\ref{tab:slotratio}, we report the value vocabulary size (i.e., the number of candidate values) of each slot in MultiWOZ 2.1 \& 2.4, respectively. We also report their value change ratios. As can be observed, for some slots, the value vocabulary size decreases due to value normalization and error correction. For some slots, the value vocabulary size increases mainly because a few labels that contain multiple values have been additionally introduced. 
Table~\ref{tab:slotratio} also indicates that 
the \verb|name|-related slots have the highest  value change ratio. Since these slots usually have ``longer'' values, the annotators are more likely to make incomplete and inconsistent annotations.

\begin{table}[h]
\centering
\setlength{\tabcolsep}{4.0pt}
\begin{tabular}{l|cc|cc}
\hline
\textbf{Slot}  & \textbf{2.1} & \textbf{2.4} & \textbf{Val($\%$)}  & \textbf{Test($\%$)} \\ \hline
attraction-area        & 7   & 8   & 1.97 & 1.93 \\
attraction-name        & 106 & 92  & 5.34 & 5.16 \\
attraction-type        & 17  & 23  & 4.62 & 3.77 \\
hotel-area             & 7   & 8   & 3.92 & 3.99 \\
hotel-book day         & 8   & 8   & 0.33 & 0.52 \\
hotel-book people      & 9   & 9   & 0.68 & 0.53 \\
hotel-book stay        & 6   & 7   & 0.42 & 0.42 \\
hotel-internet         & 5   & 4   & 2.32 & 2.24 \\
hotel-name             & 48  & 46  & 6.28 & 3.95 \\
hotel-parking          & 5   & 4   & 2.54 & 2.35 \\
hotel-pricerange       & 6   & 6   & 1.76 & 2.06 \\
hotel-stars            & 8   & 10  & 1.52 & 1.44 \\
hotel-type             & 5   & 4   & 5.06 & 4.78 \\
rest.-area        & 7   & 8   & 2.18 & 2.38 \\
rest.-book day    & 8   & 11  & 0.35 & 0.27 \\
rest.-book people & 9   & 9   & 0.37 & 0.45 \\
rest.-book time   & 59  & 62  & 0.56 & 0.46 \\
rest.-food        & 89  & 93  & 2.58 & 2.28 \\
rest.-name        & 135 & 121 & 7.81 & 5.90 \\
rest.-pricerange  & 5   & 7   & 1.51 & 2.05 \\
taxi-arriveby          & 62  & 61  & 0.41 & 0.56 \\
taxi-departure         & 177 & 172 & 0.92 & 0.86 \\
taxi-destination       & 185 & 181 & 1.14 & 0.75 \\
taxi-leaveat           & 92  & 89  & 0.84 & 0.45 \\
train-arriveby         & 109 & 73  & 1.40 & 2.86 \\
train-book people      & 11  & 12  & 1.22 & 1.76 \\
train-day              & 8   & 9   & 0.31 & 0.24 \\
train-departure        & 19  & 15  & 0.71 & 1.10 \\
train-destination      & 20  & 17  & 0.71 & 1.00 \\
train-leaveat          & 128 & 96  & 4.64 & 5.12 \\ 
\hline
\end{tabular}
\caption{The slot value vocabulary size counted on the validation set and test set of MultiWOZ 2.1 and MultiWOZ 2.4, respectively, and the slot-specific value change ratio. ``rest.'' is the abbreviation of restaurant.}
\label{tab:slotratio}
\vspace*{-0.5cm}
\end{table}

\section{Per-Slot (Slot-Specific) Accuracy}

\begin{table*}[!t]
\centering
\begin{tabular}{l|ccccc}
\hline
\textbf{Slot} &
  \begin{tabular}[c]{@{}c@{}}\textbf{MultiWOZ}\\ \textbf{2.1}\end{tabular} &
  \begin{tabular}[c]{@{}c@{}}\textbf{MultiWOZ}\\ \textbf{2.2}\end{tabular} &
  \begin{tabular}[c]{@{}c@{}}\textbf{MultiWOZ}\\ \textbf{2.3}\end{tabular} &
  \begin{tabular}[c]{@{}c@{}}\textbf{MultiWOZ}\\ \textbf{2.3-cof}\end{tabular} &
  \begin{tabular}[c]{@{}c@{}}\textbf{MultiWOZ}\\ \textbf{2.4}\end{tabular} \\ \hline
attraction-area        & 95.94 & 95.97 & 96.28          & \textbf{96.80} & 96.38          \\
attraction-name        & 93.64 & 93.92 & 95.28          & 94.59          & \textbf{96.38} \\
attraction-type        & 96.76 & 97.12 & 96.53          & 96.91          & \textbf{98.24} \\
hotel-area             & 94.33 & 94.44 & 94.65          & 95.02          & \textbf{96.16} \\
hotel-book day         & 98.87 & 99.06 & 99.04          & 99.32          & \textbf{99.52} \\
hotel-book people      & 98.66 & 98.72 & 98.93          & 99.17          & \textbf{99.19} \\
hotel-book stay        & 99.23 & 99.50 & 99.70          & 99.70          & \textbf{99.88} \\
hotel-internet         & 97.02 & 97.02 & 97.45          & 97.56          & \textbf{97.96} \\
hotel-name             & 94.67 & 93.76 & 94.71          & 94.71          & \textbf{96.92} \\
hotel-parking          & 97.04 & 97.19 & 97.90          & 98.34          & \textbf{98.68} \\
hotel-pricerange       & 96.00 & 96.23 & 95.90          & 96.40          & \textbf{96.59} \\
hotel-stars            & 97.88 & 97.95 & 97.99          & 98.09          & \textbf{99.16} \\
hotel-type             & 94.67 & 94.22 & \textbf{95.92} & 95.65          & 94.75          \\
restaurant-area        & 96.30 & 95.47 & 95.52          & 96.05          & \textbf{97.52} \\
restaurant-book day    & 98.90 & 98.91 & 98.83          & \textbf{99.66} & 98.59          \\
restaurant-book people & 98.91 & 98.98 & 99.17          & 99.21          & \textbf{99.31} \\
restaurant-book time   & 99.43 & 99.24 & 99.31          & \textbf{99.46} & 99.28          \\
restaurant-food        & 97.69 & 97.61 & 97.49          & 97.64          & \textbf{98.71} \\
restaurant-name        & 92.71 & 93.18 & 95.10          & 94.91          & \textbf{96.01} \\
restaurant-pricerange  & 95.36 & 95.65 & 95.75          & 96.26          & \textbf{96.59} \\
taxi-arriveby          & 98.36 & 98.03 & 98.18          & \textbf{98.45} & 98.17          \\
taxi-departure         & 96.13 & 96.35 & 96.15          & \textbf{97.49} & 96.55          \\
taxi-destination       & 95.70 & 95.50 & 95.56          & \textbf{97.59} & 95.68          \\
taxi-leaveat           & 98.91 & 98.96 & \textbf{99.04} & 99.02          & 98.72          \\
train-arriveby         & 96.40 & 96.40 & 96.54          & 96.76          & \textbf{98.85} \\
train-book people      & 97.26 & 97.04 & 97.29          & 97.67          & \textbf{98.62} \\
train-day              & 98.63 & 98.60 & 99.04          & \textbf{99.38} & 98.94          \\
train-departure        & 98.43 & 98.40 & 97.56          & 97.50          & \textbf{99.32} \\
train-destination      & 98.55 & 98.30 & 97.96          & 97.86          & \textbf{99.43} \\
train-leaveat          & 93.64 & 94.14 & 93.98          & 93.96          & \textbf{96.96} \\ \hline
\end{tabular}
\caption{Per-slot (slot-specific) accuracy ($\%$) of SUMBT on different versions of the MultiWOZ dataset. The results on MultiWOZ 2.1-2.3 and MultiWOZ 2.3-cof are from \cite{han2020multiwoz}. It is shown that most slots demonstrate stronger performance on MultiWOZ 2.4 than on all the other versions.}
\label{tab:slotacc}
\vspace*{-0.05cm}
\end{table*}

In Section~\ref{sec:bbbb}, we have presented the joint goal accuracy and average slot accuracy of eight state-of-the-art dialogue state tracking models. The results have demonstrated that much better performance can be achieved on our refined annotations in terms of the two metrics. Here, we further report the per-slot (slot-specific) accuracy of SUMBT on different versions of the MultiWOZ dataset. The slot-specific accuracy is defined as the ratio of dialogue turns in which the value of a particular slot has been correctly predicted. The results are shown in  Table~\ref{tab:slotacc}, from which we can observe that the majority of slots (21 out of 30) demonstrate higher accuracies on MultiWOZ 2.4. Even though MultiWOZ 2.3-cof additionally introduces the co-reference annotations as a kind of auxiliary information, it still only shows the best performance in 7 slots. Compared with MultiWOZ 2.1, SUMBT has achieved higher slot-specific accuracies in 26 slots on MultiWOZ 2.4. These results confirm again the utility and validity of our refined version MultiWOZ 2.4.

\section{Case Study}

% \usepackage{multirow}
\begin{table*}[t]
\renewcommand{\arraystretch}{1.35}
\small
\centering
\begin{tabular}{c|cccc}
\hline
\textbf{Dialogue ID}               & \multicolumn{4}{c}{\textbf{Dialogue Context, Groundtruth Annotations, and Predictions of SOM-DST and STAR}}                                                                                                                                                                                                                                                                                                                                                    \\ \hline
\multirow{3}{*}{PMUL1931} & \multicolumn{4}{l}{\begin{tabular}[c]{@{}l@{}}\textbf{Sys:} We have 6 different guest houses that fit your criteria. Do you have a specific price range in mind?\\ \textbf{Usr:} No, it does not matter.\end{tabular}}                                                                                                                                                                                         \\ \cline{2-5} 
                          & \multicolumn{1}{c|}{{MultiWOZ 2.1}}                                                    & \multicolumn{1}{c|}{{MultiWOZ 2.4}}                                                           & \multicolumn{1}{c|}{{SOM-DST}}                                                                & {STAR}                                                                   \\ \cline{2-5} 
                          & \multicolumn{1}{c|}{\textit{\begin{tabular}[c]{@{}c@{}}hotel-pricerange\\ none\end{tabular}}}         & \multicolumn{1}{c|}{\textit{\begin{tabular}[c]{@{}c@{}}hotel-pricerange\\ dontcare\end{tabular}}}             & \multicolumn{1}{c|}{\textit{\begin{tabular}[c]{@{}c@{}}hotel-pricerange\\ dontcare\end{tabular}}}             & \textit{\begin{tabular}[c]{@{}c@{}}hotel-pricerange\\ dontcare\end{tabular}}             \\ \hline
\multirow{3}{*}{PMUL3158} & \multicolumn{4}{l}{\textbf{Usr:} I want to find a place in town to visit called jesus green outdoor pool.}                                                                                                                                                                                                                                                                                            \\ \cline{2-5} 
                          & \multicolumn{1}{c|}{{MultiWOZ 2.1}}                                                    & \multicolumn{1}{c|}{{MultiWOZ 2.4}}                                                           & \multicolumn{1}{c|}{{SOM-DST}}                                                                & {STAR}                                                                   \\ \cline{2-5} 
                          & \multicolumn{1}{c|}{\textit{\begin{tabular}[c]{@{}c@{}}attraction-type\\ swimming pool\end{tabular}}} & \multicolumn{1}{c|}{\textit{\begin{tabular}[c]{@{}c@{}}attraction-type\\ none\end{tabular}}}                  & \multicolumn{1}{c|}{\textit{\begin{tabular}[c]{@{}c@{}}attraction-type\\ none\end{tabular}}}                  & \textit{\begin{tabular}[c]{@{}c@{}}attraction-type\\ none\end{tabular}}                  \\ \hline
\multirow{3}{*}{MUL1489}  & \multicolumn{4}{l}{\begin{tabular}[c]{@{}l@{}}\textbf{Sys:} Ok, you are all set for cote on Friday, table for 8 at 17:30. Can I help with anything else?\\ \textbf{Usr:} Can I have the reference number for the reservation please?\\ \textbf{Sys:} Booking was unsuccessful. Can you try another time slot?\\ \textbf{Usr:} What about 16:30?\end{tabular}}                                                                    \\ \cline{2-5} 
                          & \multicolumn{1}{c|}{{MultiWOZ 2.1}}                                                    & \multicolumn{1}{c|}{{MultiWOZ 2.4}}                                                           & \multicolumn{1}{c|}{{SOM-DST}}                                                                & {STAR}                                                                   \\ \cline{2-5} 
                          & \multicolumn{1}{c|}{\textit{\begin{tabular}[c]{@{}c@{}}restaurant-book time\\ 17:30\end{tabular}}}    & \multicolumn{1}{c|}{\textit{\begin{tabular}[c]{@{}c@{}}restaurant-book time\\ 16:30\end{tabular}}}            & \multicolumn{1}{c|}{\textit{\begin{tabular}[c]{@{}c@{}}restaurant-book time\\ 16:30\end{tabular}}}            & \textit{\begin{tabular}[c]{@{}c@{}}restaurant-book time\\ 16:30\end{tabular}}            \\ \hline
\multirow{3}{*}{PMUL0550} & \multicolumn{4}{l}{\begin{tabular}[c]{@{}l@{}}\textbf{Sys:} I recommend Charlie Chan. Would you like to reserve a table?\\ \textbf{Usr:} Yes. Monday, 8 people, 10:30.\end{tabular}}                                                                                                                                                                                                                           \\ \cline{2-5} 
                          & \multicolumn{1}{c|}{{MultiWOZ 2.1}}                                                    & \multicolumn{1}{c|}{{MultiWOZ 2.4}}                                                           & \multicolumn{1}{c|}{{SOM-DST}}                                                                & {STAR}                                                                   \\ \cline{2-5} 
                          & \multicolumn{1}{c|}{\textit{\begin{tabular}[c]{@{}c@{}}restaurant-name\\ Charlie\end{tabular}}}       & \multicolumn{1}{c|}{\textit{\begin{tabular}[c]{@{}c@{}}restaurant-name\\ Charlie Chan\end{tabular}}} & \multicolumn{1}{c|}{\textit{\begin{tabular}[c]{@{}c@{}}restaurant-name\\ Charlie Chan\end{tabular}}} & \textit{\begin{tabular}[c]{@{}c@{}}restaurant-name\\ Charlie Chan\end{tabular}} \\ \hline
\multirow{3}{*}{MUL1697}  & \multicolumn{4}{l}{\begin{tabular}[c]{@{}l@{}}\textbf{Sys:} I am sorry none of them have booking available for that time, another time maybe?\\ \textbf{Usr:} Is 09:45 an available time?\end{tabular}}                                                                                                                                                                                                        \\ \cline{2-5} 
                          & \multicolumn{1}{c|}{{MultiWOZ 2.1}}                                                    & \multicolumn{1}{c|}{{MultiWOZ 2.4}}                                                           & \multicolumn{1}{c|}{{SOM-DST}}                                                                & {STAR}                                                                   \\ \cline{2-5} 
                          & \multicolumn{1}{c|}{\textit{\begin{tabular}[c]{@{}c@{}}restaurant-book time\\ 21:45\end{tabular}}}    & \multicolumn{1}{c|}{\textit{\begin{tabular}[c]{@{}c@{}}restaurant-book time\\ 09:45\end{tabular}}}            & \multicolumn{1}{c|}{\textit{\begin{tabular}[c]{@{}c@{}}restaurant-book time\\ 10:45\end{tabular}}}            & \textit{\begin{tabular}[c]{@{}c@{}}restaurant-book time\\ 09:45\end{tabular}}            \\ \hline
\end{tabular}
\caption{Examples of test set dialogues in which the annotations of MultiWOZ 2.1 are incorrect but the predictions of SOM-DST and STAR are correct (except the prediction of SOM-DST in the last example), as the predicted slot values are consistent with the dialogue context. Given that the annotations of MultiWOZ 2.4 are consistent with the dialogue context as well, there is no doubt that higher performance can be achieved when performing evaluation on MultiWOZ 2.4. Note that only the problematic slots are presented.}
\label{tab:showcase}
\end{table*}

Except for the quantitative analyses provided in the benchmark evaluation and human evaluation, we also conduct a qualitative analysis to understand more intuitively why and how the refined annotations boost the performance of evaluation. To this end, we showcase several dialogues from the test set in Table~\ref{tab:showcase}, where we include the annotations of MultiWOZ 2.1 and MultiWOZ 2.4 and also the predictions of SOM-DST and STAR. It is easy to check that the annotations of MultiWOZ 2.1 are incorrect, while the annotations of MultiWOZ 2.4 are consistent with the dialogue context and therefore are valid. From Table~\ref{tab:showcase}, we also observe that the predictions of both SOM-DST and STAR are the same as the annotations of MultiWOZ 2.4 in the first four dialogues. In the last dialogue, the prediction of STAR is consistent with the annotation of MultiWOZ 2.4, whereas the predicted slot value of SOM-DST is different from the annotations of both MultiWOZ 2.1 and MultiWOZ 2.4. These examples show that the performance of existing dialogue state tracking models is underestimated because of the invalid annotations in MultiWOZ 2.1. While MultiWOZ 2.4 can better manifest the true model performance owing to the refined annotations that align well with the dialogue context.

% Considering that MultiWOZ 2.1 and MultiWOZ 2.4 share the same training set, we can conclude that the improved performance refined annotations.

\section{Discussion}

Recall that in MultiWOZ 2.4, we only refined the annotations of the validation set and test set. The annotations in the training set remain unchanged (the same as MultiWOZ 2.1). As a result, all the benchmark models are retrained on the original noisy training set. The only difference is that we use the cleaned validation set to choose the best model and then report the results on the cleaned test set. Even so, we have shown in our empirical study that the benchmark models can obtain better performance on MultiWOZ 2.4 than on all the previous versions. Considering that all the previous refined versions also corrected the (partial) annotation errors in the training set, the superiority of MultiWOZ 2.4 indicates that existing versions have not fully resolved the incorrect and inconsistent annotations. Therefore, although there have been three refined versions, our refinement is still necessary and meaningful. In addition, the refined validation set and test set can be combined with the training set of MultiWOZ 2.3. Since MultiWOZ 2.3 has the cleanest training set by far, this combination has the potential to result in even higher performance of existing methods. 

% The cleaned validation set and test set of MultiWOZ 2.4 can more accurately reflect the true performance of existing models. 

On the other hand, it is well-understood that deep (neural) models are data-hungry. However, it is costly and labor-intensive to collect high-quality large-scale datasets, especially dialogue datasets that involve multiple domains and multiple turns. The dataset composed of a large noisy training set and a small clean validation set and test set is more common in practice. In view of this, our refined dataset is a better reflection of the realistic situation we encounter in our daily life. Moreover, a noisy training set may motivate us to design more robust and noise-resilient training paradigms. As a matter of fact, noisy label learning~\citep{han2020survey, song2020learning} has been widely studied in the machine learning community to train robust models from noisy training data. Numerous advanced techniques have been investigated as well. We hope to see that these techniques can also be applied to the study of dialogue systems and thus accelerate the development of conversational AI.

\section{Potential Impacts}

We believe that our refined dataset MultiWOZ 2.4 would have substantial impacts in academia. First of all, the cleaned validation set and test set can help us evaluate the performance of dialogue state tracking models more properly and fairly, which is undoubtedly beneficial to the research of task-oriented dialogue systems. In addition, MultiWOZ 2.4 may also serve as a potential dataset to assist the research of noisy label learning in the machine learning community. The advantage of MultiWOZ 2.4 is that it is a  multi-label dataset with real noise in the training set. In the machine learning community, it has been recognized as a future research direction to study noisy label learning for multi-label classification~\citep{song2020learning}.

% \section{Code Links}
% % The source code of all benchmark models are publicly available.

% \noindent \textbf{SUMBT:} \url{https://github.com/SKTBrain/SUMBT}

% \noindent \textbf{STAR:} \url{https://github.com/smartyfh/DST-STAR}

% \noindent \textbf{TRADE:} \url{https://github.com/jasonwu0731/trade-dst}

% \noindent \textbf{PIN:} \url{https://github.com/BDBC-KG-NLP/PIN_EMNLP2020}

% \noindent \textbf{SOM-DST:} \url{https://github.com/clovaai/som-dst}

% \noindent \textbf{SimpleTOD:} \url{https://github.com/salesforce/simpletod}

% \noindent \textbf{SAVN:} \url{https://github.com/wyxlzsq/savn}

% \noindent \textbf{TripPy:} \url{https://gitlab.cs.uni-duesseldorf.de/general/dsml/trippy-public}

\end{document}